# Latent Tensor Factorization with Nonlinear PID Control for Missing Data Recovery in Non-Intrusive Load Monitoring

Yiran Wang, Tangtang Xie and Hao Wu

*Abstract*—Non-Intrusive Load Monitoring (NILM) has emerged as a key smart grid technology, identifying electrical device and providing detailed energy consumption data for precise demand response management. Nevertheless, NILM data suffers from missing values due to inescapable factors like sensor failure, leading to inaccuracies in non-intrusive load monitoring. A stochastic gradient descent (SGD)-based latent factorization of tensors model has proven to be effective in estimating missing data, however, it updates a latent factor solely based on the current stochastic gradient, without considering past information, which leads to slow convergence of an LFT model. To address this issue, this paper proposes a Nonlinear Proportional-integral-derivative (PID)-Incorporated Latent factorization of tensors (NPIL) model with two-fold ideas: a) rebuilding the instant learning error according to the principle of a nonlinear PID controller, thus, the past update information is efficiently incorporated into the learning scheme, and b) implementing gain parameter adaptation by utilizing particle swarm optimization (PSO) algorithm, hence, the model computational efficiency is effectively improved. Experimental results on real-world NILM datasets demonstrate that the proposed NPIL model surpasses state-of-the-art models in convergence rate and accuracy when predicting the missing NILM data.

## I. Introduction

As global energy demand grows, traditional power grids face mounting challenges, impacting efficiency and reliability. With the rapid progress of smart grids, NILM has become highly important in real-world applications. It identifies the usage of various applications by analyzing the accumulated power data from numerous grids, thereby enabling household energy management and conservation. [1], however, the NILM data inevitably have missing values due to inescapable factors, e.g., device failures, channel delays, and network congestion, which can significantly reduce monitoring and decomposition accuracy. Hence, how to accurately and fast recover missing NILM data becomes a thorny issue.

To date, researchers have explored various methods to address missing data in NILM, traditional techniques like linear interpolation, k-nearest neighbor imputation [2-4], and mean imputation, can fill data gaps but are less effective with high data loss rates, some advanced approaches, deep learning models like RNNs [5-7] and LSTMs [8],[9] and other advanced methods provide stronger modeling but require significant data and computation. Matrix factorization efficiently models low-rank power data [10-21] but falters with complex nonlinear high-order features and multi-device interactions.

In particular, LFT model based missing data recovery methods have been extensively studied [22-28], it can model the spatiotemporal incomplete data into a high-dimensional incomplete tensor [29-32], hence, the spatiotemporal information of the observed data can be preserved fully. For example, Wu et al. [33] proposed six regularized non-negative LFT models. Chen et al. [34] proposed a fast non-negative tensor decomposition model integrating the generalized momentum method and PSO. Wu et al. [35] proposed a method combining fine-grained regularization based on Swish-p and a fuzzy controller for adaptive hyperparameter tuning, additionally. Fang et al. [36] proposed a model combining nonnegative tensor RESCAL decomposition with modularity maximization. Wu et al. [37] proposed a Cauchy loss-based non-negative LFT model. Luo et al. [38] proposed a nonnegative latent-factorization of tensors model based on the alternating direction method of multipliers. Wu et al. [39] proposed four LFT-based representation learning methods for effectively extracting useful knowledge from HDI dynamic networks. Luo et al. [40] proposed a deeply adjustable non-negative tensor LFT model, these models effectively estimate missing data for HDI tensors, LFT model mostly rely on a standard SGD algorithm for optimization [41-45], however, the standard SGD algorithm updates a latent factor solely based on the current stochastic gradient, without considering past information, thereby leading to slow convergence.

Recent research shows that PID controller can enhance the efficiency of SGD [46-49]. For example, Wu et al. [49] combine PID with the LFT model, achieving competitive prediction accuracy through an adjusted force error based on PID principles. An et al. [50] proposed using PID controllers to reconstruct gradients for faster model convergence. However, a PID controllers are limited in solving nonlinear problems [51-53]. Li et al. [54] suggest incorporating a nonlinear PID controller into the learning scheme to improve the performance of latent factor analysis models, enabling rapid convergence and enhancing the model's capability to learn from incomplete data.

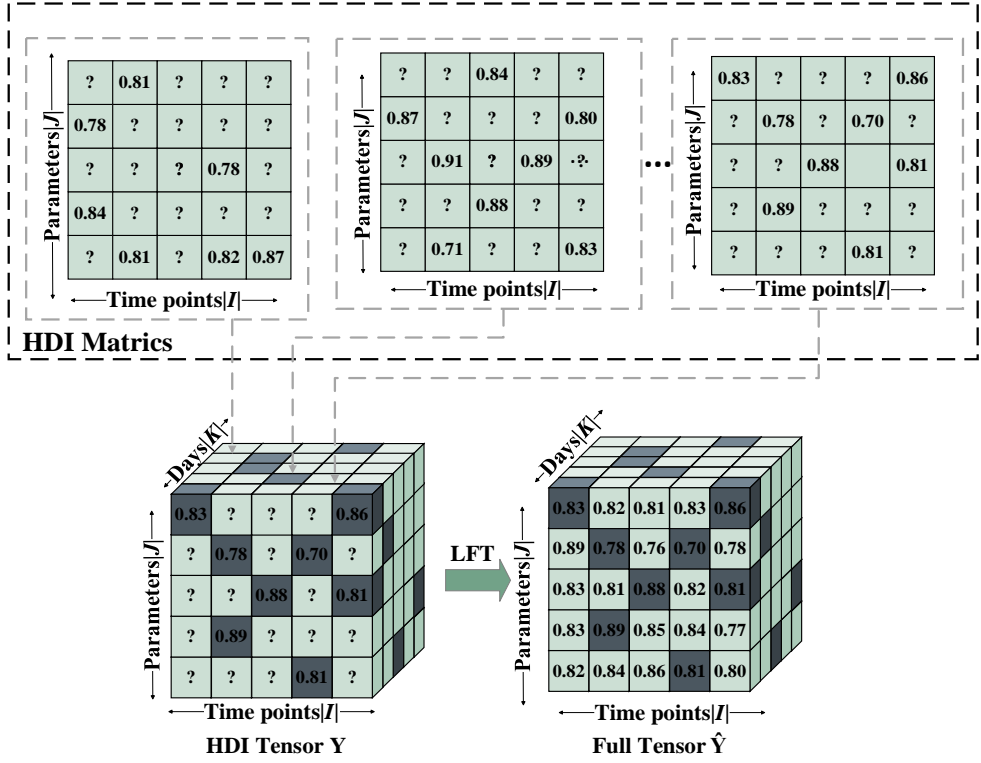

Fig. 1. NILM Data to third-order HDI Tensor

Motived by these discoveries, this paper designs a nonlinear PID-Incorporated LFT model (NPIL) for effectively and efficiently recover missing NILM data. It rebuilds the instant learning error via incorporating the past update information into the learning scheme according to the principle of a nonlinear PID controller. Moreover, the PSO algorithm is adopted to implement gain parameter adaptation. Therefore, the main contributions of this study include:

a) An NPIL model. It models the NILM data as an HDI tensor, and implements highly accurate and efficient estimation for missing NILM data.

b) A parameter self-adaptation learning scheme. The gain parameters are realized adaptively by using the PSO algorithm.

Experiments on two real NILM datasets (iAWE and UK-DALE) demonstrate that the proposed NPIL model outperforms state-of-the-art models in both estimation accuracy and time cost for missing NILM data recovery.

Section II introduces the preliminaries, Section III presents the construction process of NPIL, Section IV reports the experimental results and analyses, and Section V provides a summary of the study.

## II. PRELIMINARIES

### A. Latent Factorization of Tensor(LFT)

Given an HDI tensor **Y** with dimensions ($|I| \times |J| \times |K|$), Canonical Polyadic (CP) decomposition is one of the most widely used decomposition [55-57], the LFT model aims to find its rank-R approximation $\hat{\mathbf{Y}}$. As shown in Fig.2, we construct an approximate tensor $\hat{\mathbf{Y}}$ by summing up rank-one tensors $\mathbf{X}_r$.

$$\hat{\mathbf{Y}} = \sum_{r=1}^{R} \mathbf{X}_r \tag{1}$$

A rank-one tensor is defined as follows: if **X** can be expressed as the outer product of three LF vectors of lengths $|I|$, $|J|$, and $|K|$, then **X** is a rank-one tensor. Thus, these LF vectors are further organized into latent feature matrix $A$, $B$, and $C$ of size ($|I| \times R$, $|J| \times R$, $|K| \times R$), each entry $\hat{y}_{ijk}$ is represented as follows:

$$\hat{y}_{ijk} = \sum_{r=1}^{R} a_{ir} b_{jr} c_{kr} \tag{2}$$

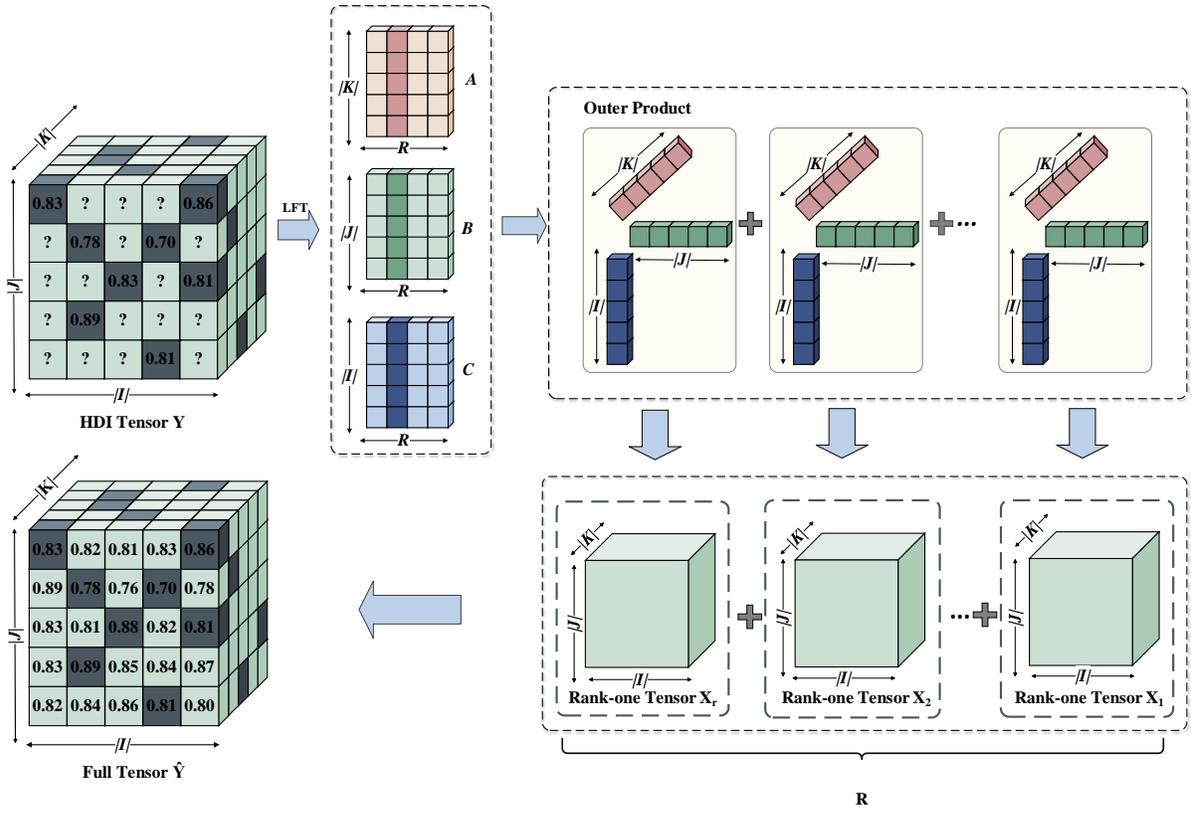

Fig. 2. Latent factorization of HDI tensor **Y**

## B. NPID Controller

The nonlinear PID controller is an extension of the PID controller, designed to handle some nonlinear systems [58]. The standard form of the NPID controller can be represented as:

$$\tilde{e}^{(t)} = K_p e^{(t)} + K_I \sum_{h=0}^{t} e^{(h)} + K_D \left(e^{(t)} - e^{(t-1)}\right) \tag{3}$$

where $t$ represents the time point of the NPID controller, hence $e^{(t)}$ represents the control error at time $t$, and $\tilde{e}^{(t)}$ is a further refinement of the error. $K_P$, $K_I$ and $K_D$ are the parameters of the proportional, integral, and derivative controllers, respectively. By applying a non-linear mapping to $K_P$, $K_I$, and $K_D$, we obtain the following expression:

$$\begin{cases} K_p = K_{p1} + K_{p2}(1 - \operatorname{sech}(K_{p3} e^{(t)})), \\ K_I = K_{i1} \operatorname{sech}(K_{i2} e^{(t)}), \\ K_D = K_{d1} + K_{d2} / (1 + K_{d3} \exp(K_{d4} e^{(t)})); \end{cases} \tag{4}$$

## C. Problem Statement

We build an objective function $\varepsilon$ to measure the difference between **Y** and $\hat{\mathbf{Y}}$. This objective function typically uses the Euclidean distance as a metric.

$$\varepsilon = \frac{1}{2} \| \mathbf{Y} - \hat{\mathbf{Y}} \|_F^2 \tag{5}$$

As shown in Fig. 1, a small example illustrates the connection between NILM data and the HDI tensor. The dynamic state of NILM data across $|K|$ time slots can be represented by a time series, with each subsequence reflecting various measurement parameters at different times throughout the day. Based on previous research [59-64], we define the loss function only on the known dataset $\Lambda$,

$$\varepsilon = \frac{1}{2} \sum_{y_{ijk} \in \Lambda} \left( y_{ijk} - \sum_{r=1}^{R} a_{ir} b_{jr} c_{kr} \right)^2 \tag{6}$$

Imbalanced target distributions can cause models to overfit common data. To prevent this and improve generalization, Tikhonov regularization can be applied to the factor matrix.

$$\varepsilon = \frac{1}{2}\sum_{y_{ijk}\in\Lambda}\left(\left(y_{ijk}-\sum_{r=1}^{R}a_{ir}b_{jr}c_{kr}\right)^2 + \lambda\sum_{r=1}^{R}\left(a_{ir}^2+b_{jr}^2+c_{kr}^2\right)\right) \quad (7)$$

Research in [65]-[68] indicates that incorporating linear biases into the learning objective improves the stability and representation learning capability of the LFT model. Linear bias can be modeled with three linear biases vectors of lengths /I/, /J/ and /K/, denoted as *u, f,* and *d*, respectively. The model with biased learning objectives is expressed as follows:

$$\varepsilon = \frac{1}{2}\sum_{y_{ijk}\in\Lambda}\left(y_{ijk}-\left(\sum_{r=1}^{R}a_{ir}b_{jr}c_{kr}+u_i+f_j+d_k\right)\right)^2$$
$$+\lambda\sum_{r=1}^{R}\left(a_{ir}^2+b_{jr}^2+c_{kr}^2\right)+\lambda\left(u_i^2+f_j^2+d_k^2\right) \quad (8)$$

Thus, the learning scheme for LF matrices and LB vectors based on SGD [69] is:

$$\arg\min_{A,B,C,u,f,d}\varepsilon \overset{SGD}{\Rightarrow} \forall i\in I, j\in J, k\in K, r\in\{1,...,R\}$$

$$\begin{cases} a_{ir} \leftarrow a_{ir}-\eta\frac{\partial\varepsilon}{\partial a_{ir}}=a_{ir}+\eta\left(e_{ijk}b_{jr}c_{kr}-\lambda a_{ir}\right) \\ b_{jr} \leftarrow b_{jr}-\eta\frac{\partial\varepsilon}{\partial b_{jr}}=b_{jr}+\eta\left(e_{ijk}a_{ir}c_{kr}-\lambda b_{jr}\right) \\ c_{kr} \leftarrow c_{kr}-\eta\frac{\partial\varepsilon}{\partial c_{kr}}=c_{kr}+\eta\left(e_{ijk}a_{ir}b_{jr}-\lambda c_{kr}\right) \\ u_i \leftarrow u_i-\eta\frac{\partial\varepsilon}{\partial u_i}=u_i+\eta\left(e_{ijk}-\lambda u_i\right) \\ f_j \leftarrow f_j-\eta\frac{\partial\varepsilon}{\partial f_j}=f_j+\eta\left(e_{ijk}-\lambda f_j\right) \\ d_k \leftarrow d_k-\eta\frac{\partial\varepsilon}{\partial d_k}=d_k+\eta(e_{ijk}-\lambda d_k) \end{cases} \quad (9)$$

where $\eta$ denotes the learning rate, $\lambda$ denotes regularization coefficient. $e_{ijk}=y_{ijk}-\hat{y}_{ijk}$ denotes an instant error.

III. THE NPIL MODEL

*A. Models*

An effective controller monitors and adjusts systems in real time to ensure stability and performance. Research indicates that refining feedback control errors accelerates system convergence. Integrating the controller into machine learning objectives optimizes adjustment errors, reducing discrepancies between predicted and actual values. Based on (3) and (4), the refinement rule for the NPID learning error can be derived to obtain the adjusted instantaneous error $\tilde{e}_{ijk}^{(t)}$.

$$\tilde{e}_{ijk}^{(t)} = \left(K_{p1}+K_{p2}\left(1-\mathrm{sech}\left(K_{p3}e_{ijk}^{(t)}\right)\right)\right)e_{ijk}^{(t)}$$
$$+K_{i1}\mathrm{sech}\,K_{i2}e_{ijk}^{(t)}\sum_{h=0}^{t}e_{ijk}^{(h)}$$
$$+\left(K_{d1}+K_{d2}/\left(1+K_{d3}\exp\left(K_{d4}e_{ijk}^{(t)}\right)\right)\right)\left(e_{ijk}^{(t)}-e_{ijk}^{(t-1)}\right) \quad (10)$$

In formula (10), $e_{ijk}^{(t)}$ represents the learning residual of the current model, which is $y_{ijk}-\hat{y}_{ijk}$. The integral term corresponds to the accumulation of errors over time, which can eliminate persistent static errors and keep the system near the target value. The derivative term predicts the future trend of the error by responding to its rate of change, suppressing rapid system responses, and reducing excessive overshoot.

Hence, by substituting $\tilde{e}_{ijk}^{(t)}$ into (9), a NPID-incorporated parameter learning scheme for desired LF matrices and LB vectors is obtained as:

$$\arg\min_{A,B,C,u,f,d} \varepsilon \overset{SGD}{\Rightarrow} \forall i \in I, j \in J, k \in K, r \in \{1,\dots,R\}$$

$$\begin{cases}
a_{ir}^{l+1} \leftarrow a_{ir}^{l} - \eta \dfrac{\partial \varepsilon}{\partial a_{ir}^{l}} \\
\quad = a_{ir}^{l} + \eta \left( \tilde{e}_{ijk}^{l} b_{jr}^{l} c_{kr}^{l} - \lambda a_{ir}^{l} \right) \\
b_{jr}^{l+1} \leftarrow b_{jr}^{l} - \eta \dfrac{\partial \varepsilon}{\partial b_{(q)jr}^{l}} \\
\quad = b_{(q)jr}^{l} + \eta \left( \tilde{e}_{ijk}^{l} a_{ir}^{l} c_{kr}^{l} - \lambda b_{jr}^{l} \right) \\
c_{kr}^{l+1} \leftarrow c_{kr}^{l} - \eta \dfrac{\partial \varepsilon}{\partial c_{kr}^{l}} \\
\quad = c_{kr}^{l} + \eta \left( \tilde{e}_{ijk}^{l} a_{ir}^{l} b_{jr}^{l} - \lambda c_{kr}^{l} \right) \\
u_i^{l+1} \leftarrow u_i^{l} - \eta \dfrac{\partial \varepsilon}{\partial u_i^{l}} = u_i^{l} + \eta \left( \tilde{e}_{ijk}^{l} - \lambda u_i^{l} \right) \\
f_j^{l+1} \leftarrow f_j^{l} - \eta \dfrac{\partial \varepsilon}{\partial f_j^{l}} = b_{jr}^{l} + \eta \left( \tilde{e}_{ijk}^{l} - \lambda f_j^{l} \right) \\
d_k^{l+1} \leftarrow d_k^{l} - \eta \dfrac{\partial \varepsilon}{\partial d_k^{l}} = d_k^{l} + \eta \left( \tilde{e}_{ijk}^{l} - \lambda d_k^{l} \right)
\end{cases} \quad (11)$$

*B. Adaptation of Gain Parameters*

In (4), adjusting gain parameters during training can be resource intensive. The PSO algorithm seeks optimal solutions by adjusting particle positions and velocities [70-73]. Initially, $Q$ particles are randomly placed in a $D$-dimensional space ($D=9$), each with position and velocity vectors.

$$\begin{cases}
s_q = \begin{bmatrix} K_{(q)p1}, K_{(q)p2}, K_{(q)p3}, K_{(q)i1}, K_{(q)i2}, \\ K_{(q)d1}, K_{(q)d2}, K_{(q)d3}, K_{(q)d4} \end{bmatrix} \\
v_q = \begin{bmatrix} V_{(q)K_{p1}}, V_{(q)K_{p2}}, V_{(q)K_{p3}}, V_{(q)K_{i1}}, V_{(q)K_{i2}} \\ V_{qK_{d1}}, V_{qK_{d2}}, V_{qK_{d3}}, V_{qK_{d4}} \end{bmatrix}
\end{cases} \quad (12)$$

The fitness function for the $q$-th particle in the $l$-th iteration is calculated using its current position.

$$I_q = \rho \sqrt{\dfrac{\sum_{g_{m,n,z} \in \Omega} \left( g_{(q)m,n,z} - \hat{g}_{(q)m,n,z} \right)^2}{|\Omega|}}$$

$$+ \mu \dfrac{\sum_{g_{m,n,z} \in \Omega} \left| g_{(q)m,n,z} - \hat{g}_{(q)m,n,z} \right|_{abs}}{|\Omega|}$$

$$I_0^l = I_Q^{l-1}, \quad F_q^l = \dfrac{I_q^l - I_{q-1}^l}{I_Q^l - I_Q^{l-1}} \quad (13)$$

where $I_q$ represents the generalized error of the $q$-th particle, $g_{(q)m,n,z}$ represents the true value of the $q$-th particle on the validation set, and $\hat{g}_{(q)m,n,z}$ represents the estimated value of the $q$-th particle in the realization, where $p=u=0.5$. Thus, the velocity and position of the $q$-th particle in the $l$-th iteration can be expressed as:

$$\begin{cases}
v_q^{(l)} = w v_q^{(l-1)} + c_1 r_1 \left( b_q^{(l-1)} - s_q^{(l-1)} \right) + c_2 r_2 \left( b_q^{(l-1)} - s_q^{(l-1)} \right) \\
s_q^{(l)} = s_q^{(l-1)} + v_q^{(l)}
\end{cases} \quad (14)$$

The velocity and position of each particle are updated using individual and global best information. This process includes an inertia weight $w$, acceleration coefficients $c_1$ and $c_2$, and random variables $r_1$ and $r_2$ uniformly distributed in [0, 1]. The following rules ensure that the velocity and position values of each particle are within the specified range:

$$\forall d \in \{1,\dots,D\}: \begin{cases} s_{q,d}^{(t)} = \min\left( \breve{s}_d, \max\left( \hat{s}_d, s_{q,d}^{(t)} \right) \right), \\ v_{q,d}^{(t)} = \min\left( v_d, \max\left( \hat{s}_d, s_{q,d}^{(t)} \right) \right), \end{cases}$$

$$\breve{v} = m \times (\breve{s} - \hat{s})$$
$$\breve{v} = -\hat{v} \quad (15)$$

Here, $\breve{s}$ and $\hat{s}$ represent the lower and upper bounds of the particle's position, respectively, while $\breve{v}$ and $\hat{v}$ represent the lower and upper bounds of the particle's velocity, respectively. *m* denotes the velocity constraint parameter.

$$a_{(q)ir}^{l+1} \leftarrow a_{(q)ir}^{l} - \eta \frac{\partial \varepsilon}{\partial a_{(q)ir}^{l}} = a_{(q)ir}^{l} + \eta \left( \tilde{e}_{(q)ijk}^{l} b_{(q)jr}^{l} c_{(q)kr}^{l} - \lambda a_{(q)ir}^{l} \right) \quad (16)$$

$$b_{(q)jr}^{l+1} \leftarrow b_{(q)jr}^{l} - \eta \frac{\partial \varepsilon}{\partial b_{(q)jr}^{l}} = b_{(q)jr}^{l} + \eta \left( \tilde{e}_{(q)ijk}^{l} a_{(q)ir}^{l} c_{(q)kr}^{l} - \lambda b_{(q)jr}^{l} \right) \quad (17)$$

$$c_{(q)kr}^{l+1} \leftarrow c_{(q)kr}^{l} - \eta \frac{\partial \varepsilon}{\partial c_{(q)kr}^{l}} = c_{(q)kr}^{l} + \eta \left( \tilde{e}_{(q)ijk}^{l} a_{(q)ir}^{l} b_{(q)jr}^{l} - \lambda c_{(q)kr}^{l} \right) \quad (18)$$

$$u_{(q)i}^{l+1} \leftarrow u_{(q)i}^{l} - \eta \frac{\partial \varepsilon}{\partial u_{(q)i}^{l}} = u_{(q)i}^{l} + \eta \left( \tilde{e}_{(q)ijk}^{l} - \lambda u_{(q)i}^{l} \right) \quad (19)$$

$$f_{(q)j}^{l+1} \leftarrow f_{(q)j}^{l} - \eta \frac{\partial \varepsilon}{\partial f_{(q)j}^{l}} = f_{(q)j}^{l} + \eta \left( \tilde{e}_{(q)ijk}^{l} - \lambda f_{(q)j}^{l} \right) \quad (20)$$

$$d_{(q)k}^{l+1} \leftarrow d_{(q)k}^{l} - \eta \frac{\partial \varepsilon}{\partial d_{(q)k}^{l}} = d_{(q)k}^{l} + \eta \left( \tilde{e}_{(q)ijk}^{l} - \lambda d_{(q)k}^{l} \right) \quad (21)$$

Where $a_{(q)ir}^{l+1}$, $b_{(q)jr}^{l+1}$, $c_{(q)kr}^{l+1}$, $u_{(q)i}^{l+1}$, $f_{(q)j}^{l+1}$ and $d_{(q)k}^{l+1}$ denotes the states of the LF matrices (i.e., *A*, *B*, and *C*) and LB vectors (i.e., ***u, f, d***) during the *L*-th evolutionary iteration. Each evolutionary iteration is associated with the *Q*-th particle, and each contains *Q* such sub-iterations. According to (12)-(15), an adaptive parameter learning scheme is obtained for NPIL model.

IV. EXPERIMENTAL RESULTS AND ANALYSIS

*A. General settings*

*1)Datasets:*

The experiments use two publicly available datasets, iAWE[81], referred to as D1, and UK-DALE[82], referred to as D2, the iAWE dataset provides 73 days of electricity consumption data from household users, including measurements from the main meter and multiple household appliance monitoring devices. Similarly, the UK-DALE dataset contains power consumption data from five residential buildings in the UK collected between 2013 and 2015. For this experiment, we selected continuous 28 days of data from the iAWE and UK-DALE datasets as experimental samples, focusing on total power, apparent power and voltage. The final dataset size per sampling point is 86400 × 3 × 28 data points, random missing scenarios with 95%, 90%, and 85% missing rates were set to validate the proposed method. For details, refer to TABLE III, we split the dataset into non- overlapping training, testing, and validation sets (*K, Ω, Ψ*) in a ratio of 8:1:1, and we set the following conditions for the training process

1) The training process of the model terminates if the number of iterations reaches a predefined threshold of 500, or it converges if the error difference between two consecutive iterations is less than $10^{-4}$, additionally, we set *R*=20.
2) The PSO-related parameters are set based on empirical values established in previous research [78-80] the number of particles is set to 5. and the learning factors $C_1$ and $C_2$ are set to 2.0, the inertia weight *w* is set to 0.729. TABLE IV records the search range of gain parameters for PSO
3) Hyper-parameter settings: we conducted a grid search for the relevant hyper-parameters on one of the 15 independent experimental runs to determine their best performance, and then we applied the same settings to the remaining 14 experimental runs. on D1, we set *η* to 0.002, *λ* to 0.01, on D2, we set *η* to 0.004, *λ* to 0.01 as well.

*2) Evaluation Metrics:*

We adopt root mean square error (RMSE) and mean absolute error (MAE) as the primary evaluation metrics. These metrics are renowned for their effectiveness in quantifying the predictive performance of machine learning models across various domains.

$$RMSE = \sqrt{\frac{\sum_{y_{ijk} \in \psi} (y_{ijk} - \hat{y}_{ijk})^2}{|\psi|}}$$

$$MAE = \frac{\sum_{y_{ijk} \in \psi} |y_{ijk} - \hat{y}_{ijk}|}{|\psi|} \quad (22)$$

The smaller MAE and RMSE mean the better accuracy.

*B. Comparison with State-of-the-art Models*

This set of experiments compares the NPIL model with several state-of-the-art models to verify its performance. Models M1-M6 were tested on an Intel Core i7-13700 CPU with 32GB of RAM, using Python as the programming language.

TABLE I. DATASET DETAILS

| Dataset | Time | Day | Parameter | Elements | Density |
|---|---|---|---|---|---|
| iAWE (D1) | 86400 | 28 | 3 | 1088640 | 15% |
|  | 86400 | 28 | 3 | 725760 | 10% |
|  | 86400 | 28 | 3 | 362880 | 5% |
| UK-DALE (D2) | 86400 | 28 | 3 | 1088640 | 15% |
|  | 86400 | 28 | 3 | 725760 | 10% |
|  | 86400 | 28 | 3 | 362880 | 5% |

TABLE II. SEARCHING RANGE OF GAIN PARAMETER

|  | $Kp1$ | $Kp1$ | $Kp3$ |
|---|---|---|---|
| Range | [2,6] | [0.001,0.5] | [0.001,0.01] |
|  | $Ki1$ | $Ki2$ | $Kd1$ |
| Range | [0.0,0.001] | [0.001,0.01] | [0.0,0.001] |
|  | $Kd2$ | $Kd3$ | $Kd4$ |
| Range | [0.001,0.01] | [0.001,0.5] | [0.001,0.01] |

TABLE III. RMSE/MAE OF MODELS WITH DIFFERENT MISSING DATA ON D1/D2

| Case |  | M1 | M2 | M3 | M4 | M5 | M6 |
|---|---|---|---|---|---|---|---|
| 15%(D1) | RMSE | $0.0299_{\pm 0.0004}$ | $0.0365_{\pm 0.0010}$ | $0.0270_{\pm 8E\text{-}05}$ | $0.0375_{\pm 0.0006}$ | $0.0342_{\pm 0.0005}$ | **$0.0256_{\pm 1E\text{-}05}$** |
|  | MAE | $0.0221_{\pm 0.0005}$ | $0.0250_{\pm 0.0011}$ | $0.0199_{\pm 5E\text{-}05}$ | $0.0258_{\pm 0.0006}$ | $0.0227_{\pm 0.0005}$ | **$0.0185_{\pm 4E\text{-}05}$** |
| 10%(D1) | RMSE | $0.0346_{\pm 0.0004}$ | $0.0371_{\pm 0.0006}$ | $0.0275_{\pm 0.0003}$ | $0.0591_{\pm 0.0020}$ | $0.0533_{\pm 0.0002}$ | **$0.0259_{\pm 2E\text{-}05}$** |
|  | MAE | $0.0236_{\pm 0.0010}$ | $0.0264_{\pm 0.0010}$ | $0.0199_{\pm 5E\text{-}05}$ | $0.0302_{\pm 0.0024}$ | $0.0249_{\pm 0.0004}$ | **$0.0190_{\pm 4E\text{-}05}$** |
| 5%(D1) | RMSE | $0.0435_{\pm 0.0003}$ | $0.0459_{\pm 0.0010}$ | $0.0312_{\pm 0.0003}$ | $0.0952_{\pm 0.0007}$ | $0.0868_{\pm 0.0006}$ | **$0.0263_{\pm 9E\text{-}05}$** |
|  | MAE | $0.0252_{\pm 0.0009}$ | $0.0253_{\pm 0.0010}$ | $0.0201_{\pm 5E\text{-}05}$ | $0.0407_{\pm 0.0014}$ | $0.0313_{\pm 0.0005}$ | **$0.0191_{\pm 67E\text{-}05}$** |
| 15%(D2) | RMSE | $0.0285_{\pm 0.0002}$ | $0.0363_{\pm 0.0008}$ | $0.0253_{\pm 8E\text{-}05}$ | $0.0373_{\pm 0.0009}$ | $0.0331_{\pm 0.0004}$ | **$0.0245_{\pm 9E\text{-}05}$** |
|  | MAE | $0.0199_{\pm 0.0003}$ | $0.0244_{\pm 0.0010}$ | $0.0184_{\pm 3E\text{-}05}$ | $0.0255_{\pm 0.0008}$ | $0.0210_{\pm 0.0004}$ | **$0.0175_{\pm 5E\text{-}05}$** |
| 10%(D2) | RMSE | $0.0331_{\pm 0.0009}$ | $0.0384_{\pm 0.0004}$ | $0.0264_{\pm 0.0010}$ | $0.0545_{\pm 0.0009}$ | $0.0488_{\pm 0.0005}$ | **$0.0247_{\pm 4E\text{-}05}$** |
|  | MAE | $0.0213_{\pm 0.0006}$ | $0.0246_{\pm 0.0006}$ | $0.0191_{\pm 4E\text{-}05}$ | $0.0284_{\pm 0.0013}$ | $0.0234_{\pm 0.0005}$ | **$0.0178_{\pm 7E\text{-}05}$** |
| 5%(D2) | RMSE | $0.0427_{\pm 0.0006}$ | $0.0457_{\pm 0.0010}$ | $0.0303_{\pm 0.0003}$ | $0.1006_{\pm 0.0034}$ | $0.0871_{\pm 0.0004}$ | **$0.0248_{\pm 4E\text{-}05}$** |
|  | MAE | $0.0240_{\pm 0.0005}$ | $0.0255_{\pm 0.0013}$ | $0.0206_{\pm 8E\text{-}05}$ | $0.0478_{\pm 0.0043}$ | $0.0306_{\pm 0.0006}$ | **$0.0182_{\pm 4E\text{-}05}$** |

TABLE IV. TIME COST(S) OF MODELS ON D1/D2

| Model | iAWE | | | UK-DALE | | |
|---|---|---|---|---|---|---|
|  | 15% | 10% | 5% | 15% | 10% | 5% |
| M1 | $24.38_{\pm 3.84}$ | $21.55_{\pm 2.95}$ | $13.33_{\pm 1.79}$ | $23.00_{\pm 3.40}$ | $18.41_{\pm 5.17}$ | $14.31_{\pm 3.66}$ |
| M2 | $438.53_{\pm 21.24}$ | $683.58_{\pm 30.78}$ | $440.94_{\pm 50.07}$ | $426.04_{\pm 15.65}$ | $459.14_{\pm 31.31}$ | $423.19_{\pm 12.43}$ |
| M3 | $229.92_{\pm 13.17}$ | $83.71_{\pm 1.54}$ | $58.41_{\pm 0.89}$ | $133.25_{\pm 17.71}$ | $118.88_{\pm 32.48}$ | $80.84_{\pm 10.66}$ |
| M4 | $38.06_{\pm 13.97}$ | $20.68_{\pm 3.31}$ | $16.00_{\pm 4.54}$ | $29.67_{\pm 3.01}$ | $18.16_{\pm 2.01}$ | $8.09_{\pm 0.64}$ |
| M5 | $13.89_{\pm 6.39}$ | $8.78_{\pm 2.32}$ | $6.04_{\pm 3.18}$ | $9.55_{\pm 1.25}$ | $6.53_{\pm 0.57}$ | $4.38_{\pm 0.90}$ |
| M6 | **$10.21_{\pm 1.81}$** | **$8.57_{\pm 1.79}$** | **$5.06_{\pm 0.59}$** | **$8.93_{\pm 2.37}$** | **$5.57_{\pm 0.87}$** | **$4.15_{\pm 0.94}$** |

**M1**: [66] proposed a non-negative tensor decomposition model, BNLFT, with linear bias.

**M2**: [74] proposed a predictive model, HDOP, based on tensor-based concepts in multilinear algebra.

**M3**: [75] proposed a CP decomposition model that updates using alternating least squares and gradient descent algorithms.

**M4**: [76] proposed a tensor decomposition method based on sparsity and graph regularization

**M5**: [77] developed a QoS prediction model resilient to outliers using Cauchy loss.

**M6**: NPIL, the model proposed in this paper.

We ran 15 independent tests for models M1-M6, using a grid search to identify the optimal hyperparameter settings for each model. Fig. 4 displays the training curves, and TABLE II shows the time costs of the models, based on these results, we can draw the following conclusions:

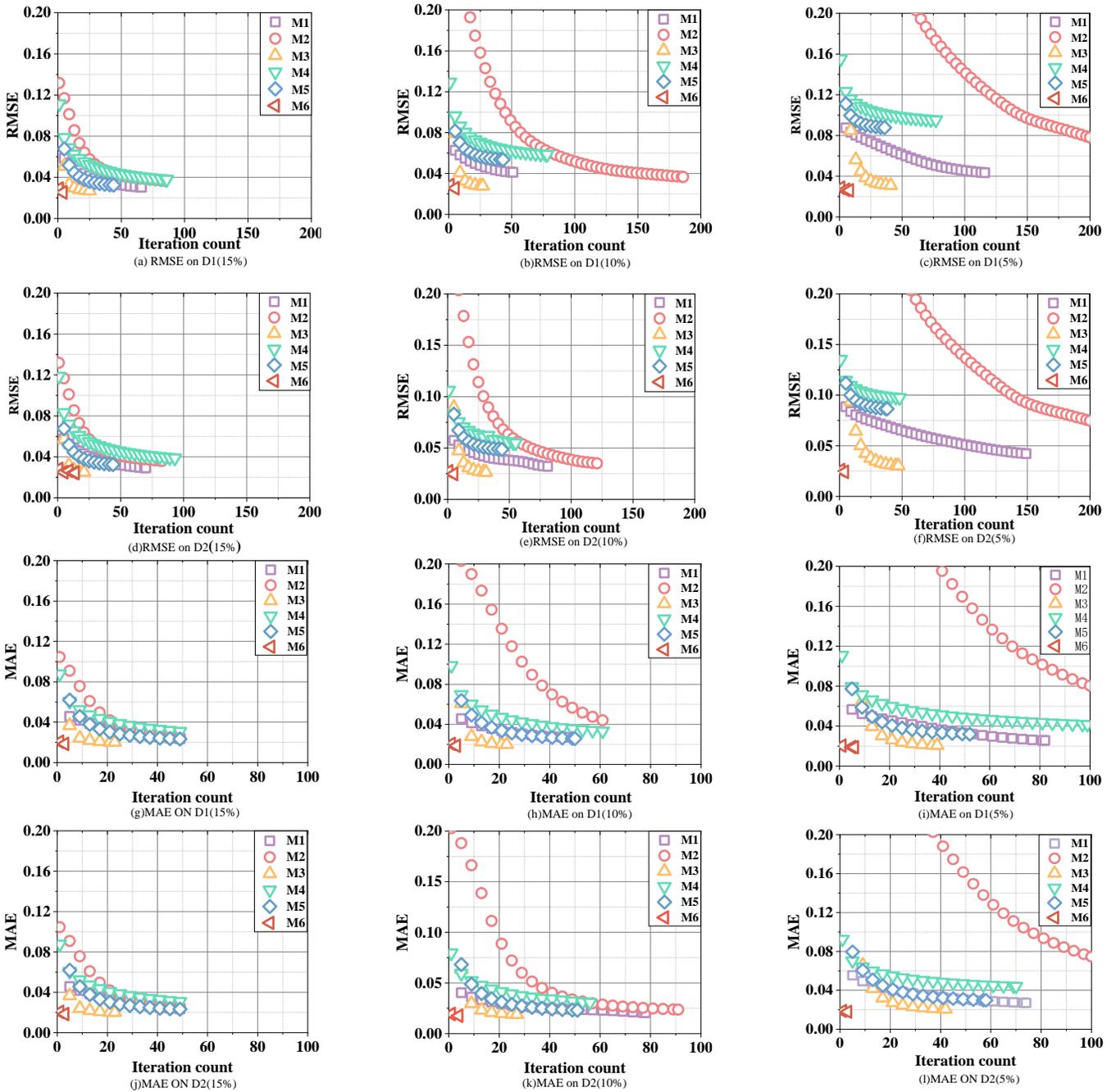

Fig. 3. Training curves of M1-M6 on D1/D2

**a) M6 demonstrates high prediction accuracy.** As shown in TABLE I, On the D1 with a density of 15%, M6 achieves RMSE of 0.0256 and MAE of 0.0185, reducing these by 31.73% and 28.29% vs. M4. Compared to M3, NPIL's RMSE and MAE are lower by 5.19% and 7.04%, respectively. Notably, as D1's data density decreased from 15% to 5%, NPIL's RMSE increased only slightly from 0.0256 to 0.0263 (2.67%), whereas M5's RMSE rose significantly from 0.0342 to 0.0868 (60.59%). A similar phenomenon occurs in D2, demonstrating NPIL's high accuracy, adaptability, sparse data handling, and efficient information use across varying data densities.

**b) M6 exhibits excellent convergence speed.** As shown in Fig. 4, M6's RMSE on D2 with a density of 15% reaches convergence in just 5 iterations, with a similar effect observed for MAE, which converges in just 3 iterations. A similar phenomenon can also be observed in D1. These results indicate that the introduction of the NPID controller is both effective and efficient. It should be noted that each iteration here includes $J$ sub-iterations to achieve the evolution of the controller gain parameters.

**c) M6 demonstrates exceptional computational efficiency.** As shown in TABLE II, On D1 with a density of 15%.M6 converges

in approximately 10.21 seconds, while M2, M3, and M4 take 438.53, 229.92, and 38.06 seconds, respectively, resulting in time cost reductions of 97.67%, 95.53%, and 73.17%, 26.49% lower than M5's 13.89s, M6 incurs significant time costs. Similar trends are observed in dataset D2, leading to the conclusion that M6 effectively reduces time costs.

**d) Summary:** We conclude that NPIL has the following characteristics: fast convergence, efficient computation, and competitive prediction accuracy. Thus, NPIL can be regarded as an effective model for power data analysis and prediction

## V. CONCLUSION

This study proposes a NILM missing data recovery method called the Nonlinear PID-Incorporated LFT model (NPIL), which integrates a NPID controller into an LFT model to adjust the model's instant learning error. The NPID controller's gain parameters are adapted through a self-adaptive parameter learning scheme based on PSO. Experimental results demonstrate the method's effectiveness in handling missing data. Future work may explore: 1) incorporating other advanced controllers for improved performance and 2) utilizing hyperparameter adaptation methods, such as fuzzy logic, for further enhancement.